\documentclass[twoside,11pt]{article}
\usepackage{color}
\usepackage{jmlr2e}
\usepackage{listings}
\usepackage[usenames,dvipsnames]{pstricks}
\usepackage[rotateright]{rotating}
\usepackage{epsfig}
\usepackage{pstricks-add}
\usepackage{pst-grad} % For gradients
\usepackage{pst-plot} % For axes
\newcommand{\minitab}[2][1]{\begin{tabular}{#1}#2\end{tabular}}
\RequirePackage[OT1]{fontenc}
\RequirePackage{hypernat}
\RequirePackage{soul}
\usepackage{amsfonts}
\usepackage{dcolumn}
\usepackage{textcomp}
\usepackage{longtable}
\usepackage{verbatim}
\usepackage{latexsym}

\usepackage{multirow}
\usepackage[linesnumbered,ruled,vlined]{algorithm2e}
\usepackage{url}
\usepackage{comment}

\usepackage{float}

\begin{document}

\title{Formulas for Counting the Sizes   of Markov Equivalence Classes  of  Directed
Acyclic Graphs}

\author{\name Yangbo He  \email heyb@pku.edu.cn \\
        \addr LMAM, School of Mathematical Sciences, LMEQF, and Center for Statistical Science,         \\ Peking University\\
         \name Bin Yu \email binyu@stat.berkeley.edu \\ \addr Departments of Statistics and EECS,          UC
        Berkeley\\
}
\ShortHeadings{Formulas for counting the sizes   of Markov Equivalence Classes}{Y.B. He}
\editor{  }
\maketitle

\begin{abstract}

 % The sizes  of    Markov equivalence classes play important roles  in both  interventional effect inferring and   causal  structure learning.
The sizes  of    Markov equivalence classes of directed acyclic graphs play important roles in measuring the  uncertainty and  complexity   in  causal learning. A Markov equivalence class can be represented   by an essential graph  and its  undirected   subgraphs determine  the size of the class.
In this paper, we  develop a method  to derive the formulas for counting the  sizes of  Markov equivalence classes. We first introduce a new concept of core graph. The size of a Markov equivalence class  of interest is a polynomial of the number of vertices given its core graph.  Then, we discuss   the recursive and explicit formula of the  polynomial,  and provide  an algorithm  to derive  the size formula via symbolic computation for any given core graph. The proposed size formula derivation sheds   light on  the relationships between the size of a Markov equivalence class and its representation graph, and makes size counting efficient,  even when   the  essential graphs     contain non-sparse undirected  subgraphs.

%Our experiments  show  that the formula-based algorithms  accelerate   size counting    dramatically  when  the  graphs   contain dense undirected  subgraphs.

\end{abstract}
\begin{keywords}
  Directed acyclic graph; Markov equivalence class;  Size formula; Causality
\end{keywords}
\section{Introduction}

A Markov Equivalence class contains all statistically equivalent  models  of  directed acyclic graphs (DAG)  \citep{pearl2000causality,spirtes2001causation}.  In general, observational data is not
sufficient to distinguish  an underlying DAG from the others in the same Markov equivalence class. The size of a   Markov
equivalence class   is the number of DAGs  in the class. It plays an important part in papers to
measure the ``uncertainty" of causal graphs or to evaluate the ``complexity" of a Markov equivalence class in  causal learning \citep{he2008active,chickering2002learning}.
For example,    \citet{he2008active}  propose several criterions, all of which are defined on  the sizes of  Markov equivalence classes, to measure  the uncertainty of causal graphs for a candidate intervention; choosing interventions by minimizing these criterions  makes helpful but expensive interventions more efficient.  \citet{maathuis2009estimating}  introduce a method  to estimate the average
causal effects of  the covariates on the response by considering  the DAGs  in the equivalence class; the size of the class determines the complexity of the  estimation.

%both inferring causal effects with observational data \citep{maathuis2009estimating,hauser2015jointly} and   learning the exact causal DAG  through further interventions \citep{he2008active}.

 An essential graph   represents a Markov equivalence class and  its undirected subgraphs  determine the size of the class \citep{andersson1997characterization}.  The size of a small Markov equivalence class can be  counted via  traversal methods that list  all DAGs in the Markov equivalence class \citep{gillispie2002size}. Recently,  \citet{he2015counting} propose a size counting algorithm  that  calculates  the size of   a Markov equivalence class via partitioning the class recursively.  In general, this method  is  efficient for  Markov equivalence classes represented by  sparse   essential graphs, but  becomes much time-consuming  when the essential graphs  contain non-sparse undirected subgraphs.

Counting graphs based on formulas is usually elegant and efficient. \citet{ robinson1973counting, robinson1977counting} provide recursive formulas  to count DAGs with a given number of vertices. \citet{steinsky2003enumeration}  develops recursive   formulas  to count Markov equivalence classes of size 1. Later, \citet{ gillispie2006formulas}  introduces recursive formulas for arbitrary size, based on all configurations of the undirected essential graphs that produce this size. However, there are few formulas available for counting the size of a given Markov equivalence class, except five formulas introduced in \citet{he2015counting} for Markov equivalence classes represented by {five specific types of undirected essential graphs (trees, graphs with up to two missing edges, etc.).}

 In this paper, we focus on the   formulas for counting the size of a Markov equivalence class. We  first introduce a new concept of ``core graph", which is an undirected chordal graph  without  dominating vertices. An undirected  essential graph can be represented by its core graph and the number of dominating vertices. The size of the  corresponding Markov equivalence class is a   polynomial of the number of dominating vertices given its core graph. Then we develop an  iterative method to derive  the polynomial, and give the explicit polynomials  for both several specific types of core graphs and all core graphs with up to five missing edges.   Based on symbolic computation, we  introduce a size formula derivation algorithm and a formula-based size counting algorithm for general core graphs and Markov equivalence classes, respectively.
{Our experiments  show that the  proposed size formula derivation  is efficient in general and formula-based algorithm can  speedup  size counting dramatically for  the Markov equivalence classes represented by  essential graphs with non-sparse     undirected subgraphs.}

The rest of the paper is arranged as follows. In Section \ref{chap2}, we  give a brief introduction about Markov equivalence class and   size counting of Markov equivalence classes.
In Section \ref{chap3}, we  propose a  method to derive   the size formulas  and  to count the sizes of Markov  equivalence classes based on { these formulas}.
In Section \ref{chap4}, we study the size formulas and formula-based size counting of Markov equivalence classes experimentally. We conclude in Section \ref{chap5} and finally present all proofs in the Appendix.

\section{ Markov Equivalence Class and  Size Counting}
\label{chap2}

 A graph ${\cal G}$ consists of  a
vertex set $V$ and an edge set $E$.
A graph is directed (undirected) if all of its edges are directed (undirected).
A sequence of edges that connect distinct vertices in $V$, say $\{v_1,\cdots,v_k\}$,  is called a path from $v_1$ to $v_k$ if  either $v_i\to v_{i+1}$ or $v_i-v_{i+1}$ is in $E$ for $i =1,\cdots,k-1$.
  A path is \emph{partially directed} if  at least one  edge in the path is directed. A path is directed (undirected) if  all edges are directed (undirected).
A     \emph{cycle}  is a
    path from a vertex to itself.

A \emph{directed acyclic graph} (DAG) $\cal D$  is a directed graph without any directed cycle.
Let $V$ be the vertex set of $\cal D$ and $\tau$ be a subset of
$V$.   The \emph{induced subgraph} ${\cal D}_{\tau
}$ of $\cal D$ over $\tau$, is defined to be the graph whose vertex set is $\tau$ and whose edge set contains all of those edges of $\cal D$ with two end points in $\tau$.
A \emph{v-structure} is a three-vertex induced subgraph of $\cal D$ like $v_1\rightarrow v_2\leftarrow v_3$.
 A graph is called a \emph{chain graph} if it contains no partially  directed cycles. The isolated undirected subgraphs of the chain graph after removing all directed edges are the  chain components  of the chain graph.
 A   \emph{chord}   of a cycle   is an edge  that joins two nonadjacent vertices in the  cycle.
 An undirected graph is \emph{chordal}
if every cycle with four or more vertices has a
chord.

A graphical model is a probabilistic model for which a DAG denotes the conditional independencies between random variables.
%A graphical model comprises  a DAG and a joint {probability} distribution.
%In a causal graphical model,  each  directed edge of the corresponding DAG is interpreted
% as a direct causal influence between two vertices of the edge.
%   With a (causal) graphical model,
%   In general, the conditional independencies implied
%  by the joint probability distribution can be read from the DAG.
%These conditional independencies are called   Markov properties of the DAG \citep{pearl1988probabilistic}.
   \emph{A Markov
equivalence class}    is a set of DAGs that encode the same set of
conditional independencies.  Let the \emph{skeleton} of an arbitrary graph $\cal G$ be  the undirected graph with the same vertices and edges as  $\cal G$,  regardless of their directions.
\citet{verma1990equivalence} prove  that   two DAGs are \emph{Markov equivalent} if and only if
they have the same skeleton and the same v-structures. Moreover, \citet{andersson1997characterization} show  that a Markov equivalence class can be represented uniquely by an \emph{essential graph}, denoted by ${\cal C}$,  which has the same skeleton as ${\cal D}$, and  an edge
is directed in ${\cal C}$ if and only if it has the same orientation in every equivalent DAG of $\cal D$. An essential graph is a chain graph  and each of its  chain components   is an undirected and connected chordal graph (UCCG for short).

 Let Size$({\cal C})$  denote the size of the Markov equivalence class represented by $\cal C$ (size of $\cal C$ for short). Clearly, $\mbox{Size}({\cal C})=1$ if $\cal C$ is a DAG; otherwise $\cal C$ may contain  at least one chain component, denoted by ${\cal C}_{\tau_1},\ldots,{\cal C}_{\tau_k}$.
   We can  calculate the size of $\cal C$ by counting the DAGs in Markov equivalence classes represented by its chain components using the following equation \citep{he2008active,gillispie2002size}:

  \begin{equation}\label{formula0}
    \mbox{Size}({\cal C})= \prod_{i=1}^k \mbox{Size}({\cal C}_{\tau_i}).
 \end{equation}

Since each chain component is an undirected and connected chordal graph,
 to   obtain  the size  of a  Markov equivalence class,  it is sufficient to compute the size of Markov equivalence classes represented by these UCCGs  according to  Equation (\ref{formula0}). %Below, we give a brief introduction about size counting method proposed by \citet{he2015counting}.

Let  ${\cal U}$ be a UCCG,  $\tau$ be the vertex set of $\cal U$ and   $\cal D$ be a DAG in the equivalence class represented by $\cal U$. A vertex $v$ is a \emph{root} of $\cal D$ if all directed edges adjacent to $v$ are out of $v$, and  $\cal D$  is \emph{$v$-rooted} if   $v$ is a root of  $\cal D$. A \emph{$v$-rooted sub-class} of $\cal U$ is the set of  all $v$-rooted DAGs in the Markov equivalence class represented by $\cal U$. A \emph{$v$-rooted essential graph}  of   ${\cal U}$,  denoted by ${\cal U}^{(v)}$, is a graph that has the same skeleton as ${\cal  U}$, and  an edge
is directed in ${\cal U}^{(v)}$ if and only if it has the same orientation in every $v$-rooted DAG  of $\cal U$.
 \citet{he2015counting}  show  that a  {$v$-rooted sub-class} of $\cal U$  can be represented uniquely by a  $v$-rooted essential graph and a Markov equivalence class  can be partitioned into  sub-classes represented by its rooted essential graphs.

\begin{lemma}\label{calcu151}
 Let ${\cal U}$ be  a UCCG over $\tau=\{{v_i}\}_{i=1,\cdots, p}$,   ${\cal U}^{(v_i)}$ be  $v_i$-rooted essential graph, and  $ f({\cal U}^{(v_i)})$ be the size of $v_i$-rooted sub-class represented by ${\cal U}^{(v_i)}$.
We have $\emph{Size}({\cal U}^{(v_i)})\geq 1$ for any $i=1,\cdots, p$, and
 \begin{equation}\label{formula151}
     \emph{Size}({\cal U})= \sum_{i=1}^p  \emph{Size}({\cal U}^{(v_i)}).
 \end{equation}

\end{lemma}

\begin{comment}
For some specific types of  UCCGs, we can obtain the sizes of the corresponding Markov equivalence classes  directly \citep{he2015counting}.

 \begin{proposition}
\label{theocountall}
Let  $\mathcal{U}_{p,n}$ be a UCCG with  $p$ vertices and $n$ edges. In the following five cases, the size of the Markov equivalence class represented by ${\cal U}_{p,n}$  is determined by $p$.
\begin{enumerate}
  \item  $f({\cal U}_{p,n})=p$    for  $n=p-1$;
  \item    $f({\cal U}_{p,n})=2p$ for $n=p$;
  \item   $f(\mathcal{U}_{p,n})=(p^2-p-4)(p-3)!$   for $n=p(p-1)/2-2$;
   \item $f(\mathcal{U}_{p,n})=2(p-1)!-(p-2)! $  for   $n=p(p-1)/2-1$;
 \item    $f({\cal U}_{p,n})=p!$  for  $n=p(p-1)/2$.
\end{enumerate}

\end{proposition}

\end{comment}

For  any $i\in \{1,\cdots, p\}$, the undirected subgraphs of  ${\cal U}^{(v_i)}$ in Lemma \ref{calcu151} are UCCGs, so we   can  calculate  $\mbox{Size}({\cal U}^{(v_i)})$  in Equation (\ref{formula151}) using   Equation (\ref{formula0}).
As a result,   using Equation (\ref{formula0}) and Equation (\ref{formula151}),  {\citet{he2015counting} propose to} calculate the size of a Markov equivalence class by partitioning it  recursively into rooted sub-classes until the sizes of all these sub-classes can be completely determined by the numbers of vertices
and edges. However, when the UCCGs   contain   non-sparse subgraphs, this method  might be much time-consuming.

 In the next section, we will show that the size of the Markov equivalence class represented by a UCCG  depends on a  subgraph  of the UCCG, and introduce a size formula derivation algorithm and a formula-based counting algorithm, which can greatly accelerate size counting of Markov equivalence classes with non-sparse undirected subgraphs.

\section{Formulas for sizes of Markov equivalence classes}\label{chap3}

In this section,  we   introduce the concept of  core graph  that determines  the size formula  of a  Markov equivalence class in Section \ref{coresec}. {Then,  we discuss the recursive and explicit    formulas for the size of a Markov equivalence class given its core graph in Section \ref{formulasec}. Finally, in Section \ref{algorithmsec}, we  provide   algorithms   to derive  size formulas   and to   count the sizes of  Markov equivalence classes based on these formulas.}

 \subsection{Core graph}\label{coresec}

  A  vertex is \emph{dominating} in a UCCG $\cal U$   if it is adjacent to all other vertices in $\cal U$.  A \emph{dominating vertex pruned}   subgraph of $\cal U$   is   obtained by removing some dominating vertices   from $\cal U$. We denote  a dominating vertex pruned subgraph of $\cal U$ as   ${\cal U}^{m-}$ if it is obtained by removing  $m$ dominating vertices  from $\cal U$.  An \emph{extended} graph of $H$, denoted by ${H}^{m+}$,   is a graph obtained by   adding $m$ dominating vertices to $H$.   %The \emph{core graph} of $\cal U$ is defined  as follows.

\begin{definition}[Core graph of a UCCG]\label{coreg} The core graph of $\cal U$ is the minimal dominating vertex pruned  subgraph of $\cal U$.
\end{definition}

Let $m$ be the number of dominating vertices in  $\cal U$, $\cal K$ be  the core graph of $\cal U$.  Clearly, $\cal K$ is the same as ${\cal U}^{m-}$. If $\cal U$ is a  completed graph, all  vertices in $\cal U$  are  dominating, so the  core graph of $\cal U$ is a \emph{null graph}. Let $\cal K$ be an undirected graph over $V$.  Clearly, according to Definition \ref{coreg},
the undirected graph $\cal K$ is a core graph of some UCCG  if and only if  $\cal K$ is an undirected chordal graph without dominating vertices.
The \emph{complement} of $\cal K$, denoted by  $ {\cal K}^c$, is a graph on the same vertices and   an edge appears  in $ {\cal K}^c$  if and only if it does not occur  in $\cal K$.  Proposition \ref{corepro} presents a  property of the complement of a core graph.

\begin{proposition}[Complement of core graph]\label{corepro} Let $\cal U$ be a UCCG, $m$ be the number of dominating vertices in  $\cal U$, $\cal K$ be the core graph of $\cal U$,  $ {\cal K}^c$ be the complement of $\cal K$. We have that
  $ {\cal K}^c$ be a connected graph, and for any two edges   in $ {\cal K}^c$, either they share a common   vertex, or they are connected by an edge.
\end{proposition}

  This property  helps us to construct a core graph. In Table \ref{edgemissthree}, we list all core graphs and the corresponding complement graphs of the UCCGs with up to three missing edges.

\begin{center}
\begin{table}[h]
%[tbp]
\unitlength=0.5mm
\begin{center}
\begin{tabular}{c|c|c|c|ccc}
  \hline
 \minitab[c]{ Number\\{( missing edges)}}&0& 1 & 2 & \multicolumn{3}{c}{3} \\\hline
   $\cal K$&${\cal K}_\emptyset$   &\unitlength=0.4mm \begin{tabular}{c} \begin{picture}(25,20)(0,0) \thicklines
\put(0,10){\circle*{3}} \put(20,10){\circle*{3}}
 \end{picture} \end{tabular} &\unitlength=0.4mm \begin{tabular}{c} \begin{picture}(25,20)(0,0) \thicklines
\put(0,0){\circle*{3}} \put(20,0){\circle*{3}}
\put(10,15){\circle*{3}}
\put(0,0){\line(1,0){20}}
\end{picture} \end{tabular}&\unitlength=0.4mm \begin{tabular}{c} \begin{picture}(25,20)(0,0) \thicklines
\put(0,0){\circle*{3}} \put(20,0){\circle*{3}}
\put(10,15){\circle*{3}}
\end{picture} \end{tabular} & \unitlength=0.4mm \begin{tabular}{c} \begin{picture}(25,20)(0,0) \thicklines
\put(0,0){\circle*{3}} \put(20,0){\circle*{3}}
\put(10,15){\circle*{3}} \put(30,15){\circle*{3}}
 \put(10,15){\line(1,0){20}}
\put(0,0){\line(1,0){20}} \put(0,0){\line(2,1){30}}
\end{picture} \end{tabular} & \unitlength=0.3mm
   \begin{tabular}{c} \begin{picture}(25,35)(0,0) \thicklines
\put(0,0){\circle*{3}} \put(20,0){\circle*{3}}
\put(10,15){\circle*{3}} \put(10,30){\circle*{3}}

\put(0,0){\line(1,0){20}} \put(0,0){\line(1,3){10}} \put(20,0){\line(-1,3){10}}
\end{picture} \end{tabular}\\\hline

  $ {\cal K}^c$&  ${\cal K}_\emptyset$  &\unitlength=0.4mm \begin{tabular}{c} \begin{picture}(25,20)(0,0) \thicklines
\put(0,10){\circle*{3}} \put(20,10){\circle*{3}}
 \put(0,10){\line(1,0){20}}
\end{picture} \end{tabular} &\unitlength=0.4mm \begin{tabular}{c} \begin{picture}(25,20)(0,0) \thicklines
\put(0,0){\circle*{3}} \put(20,0){\circle*{3}}
\put(10,15){\circle*{3}}
\put(0,0){\line(2,3){10}} \put(20,0){\line(-2,3){10}}
\end{picture} \end{tabular}&\unitlength=0.4mm \begin{tabular}{c} \begin{picture}(25,20)(0,0) \thicklines
\put(0,0){\circle*{3}} \put(20,0){\circle*{3}}
\put(10,15){\circle*{3}}
 \put(0,0){\line(1,0){20}}
\put(0,0){\line(2,3){10}} \put(20,0){\line(-2,3){10}}
\end{picture} \end{tabular} & \unitlength=0.4mm \begin{tabular}{c} \begin{picture}(25,20)(0,0) \thicklines
\put(0,0){\circle*{3}} \put(20,0){\circle*{3}}
\put(10,15){\circle*{3}} \put(30,15){\circle*{3}}
 \put(20,0){\line(2,3){10}}
\put(0,0){\line(2,3){10}} \put(20,0){\line(-2,3){10}}
\end{picture} \end{tabular} & \unitlength=0.3mm
   \begin{tabular}{c} \begin{picture}(25,35)(0,0) \thicklines
\put(0,0){\circle*{3}} \put(20,0){\circle*{3}}
\put(10,15){\circle*{3}} \put(10,30){\circle*{3}}

 \put(10,15){\line(0,1){15}}
\put(0,0){\line(2,3){10}} \put(20,0){\line(-2,3){10}}
\end{picture} \end{tabular}\\\hline
\end{tabular}
\caption{Core graphs and their complements   when at most three edges are missing, $\cal K$, $ {\cal K}^c $, and ${\cal K}_\emptyset$   denote  a  core graph,  the   complement  of $\cal K$, and a null graph, respectively.}
\label{edgemissthree}
\end{center}
\end{table}

\end{center}

 Let $\cal U$ be a UCCG with  $m$ dominating vertices, $\cal K$ be the core graph of  $\cal U$. As an extended graph of $\cal K$, $\cal U$ is the same as ${\cal K}^{m+}$ regardless  the labels of vertices, so we have $\mbox{Size}({\cal U})=\mbox{Size}({\cal K}^{m+})$.
 Clearly, the size of the Markov equivalence class represented by a  UCCG $\cal U$ is determined by its core graph $\cal K$ and the number of dominating vertices $m$. For an undirected chordal graph $\cal K$ and a  nonnegative integer $m$, we define a function $f({\cal K},m)$  as following,
\begin{equation} \label{sizecore} f({\cal K},m):= \mbox{Size}({\cal K}^{m+}).
\end{equation}
%Given the core graph $\cal K$, the  $f({\cal K},m)$ is a function of $m$.

From  the   definition of the formula $ f({\cal K},m)$, we have the following lemma directly.

\begin{lemma}\label{domk} Let ${\cal K}$ be an undirected chordal graph, and  ${\cal K}^{k+}$ be an extended graph of ${\cal K}$, we have $f({{\cal K}^{k+}},m)=f({\cal K},m+k)$.
\end{lemma}

Consider the UCCGs with at most two missing edges, as shown in  Table \ref{edgemissthree},   there is only one core graph exists, so the sizes of the corresponding Markov equivalence classes are determined given the number of vertices in the UCCGs. When three edges are missing in the UCCGs, there are three  core graphs exists, so three sizes are possible given the number of vertices.   This explains the results introduced in \citet{he2015counting} that the size of a  Markov equivalence class  is determined  given the number of vertices ($p$)  only when  no more than   two edges are missing in UCCGs. %\citep{he2015counting}

The size of $\cal U$  might be very huge; for a UCCG $\cal U$ with $p$ vertices, $\mbox{Size}({\cal U})$ reaches the maximum $p!$ when $\cal U$ is a completed graph.  In general, more edges in the UCCG (more denser), more larger the corresponding class and more time-consuming of size counting.   Fortunately, a dense UCCG $\cal U$ might has  sparse core graph $\cal K$ when many dominating vertices exist. In the next section, given the core graph $\cal K$, we will discuss the formula of $f({\cal K},m)$ that can be used to speedup the enumeration of $\mbox{Size}({\cal U})$.

\subsection{Size formulas based on core graphs}\label{formulasec}

%Let $\cal K$ be an undirected chordal graph, ${\cal K}^{m+}$ be an  extended graph of $\cal K$ for any non-negative integer. The size of the Markov equivalence class represented by ${\cal K}^{m+}$ is determined by the graph $\cal K$ and the integer; that is, $ \mbox{Size}({\cal K}^{m+})=f($\cal K$,m)$, where $f($\cal K$,m)$ is a function of $\cal K$ and $m$.

In this section, we propose a method  to derive  the size formula $f({\cal K},m)$  defined in Equation (\ref{sizecore}). We  first introduce  a recursive formula of $f({\cal K},m)$ given  $\cal K$, then propose a method to derive  the explicit size formulas,  and finally    give the explicit formulas  for both several specific types of core graphs and all core graphs with up to five missing edges.

 Theorem \ref{recursiveformula1} introduces the main recursive formula for the size of a Markov equivalence class whose representation graph is extended from  an undirected chordal graph  $\cal K$ as follows.

\begin{theorem}\label{recursiveformula1}
Let $\cal K$ be an undirected chordal graph   over $V$. For  any integer $m\geq 0$,  ${\cal K}^{m+}$ is an  extended graph of $\cal K$, and   $f({\cal K},{m})$ is the size of  ${\cal K}^{m+}$ defined in Equation (\ref{sizecore}). We have $f\left({\cal K},{0}\right)=\emph{Size}({\cal K})$, and for  any integer $m>0$,
\begin{equation}\label{iterizeformul}
   f\left({\cal K},{m}\right)=  m \cdot f\left({\cal K},{m-1}\right)+\sum_{v\in V } f\left( {\cal K}_{N_v},{m}\right)  \frac{\emph{Size}\left({\cal K}^{(v)}\right)}{\emph{Size} \left({\cal K}_{N_v}\right)},
\end{equation}
where   ${\cal K}^{(v)}$ is a v-rooted graph of ${\cal K}$ and ${\cal K}_{N_v}$ is an induced subgraph   on the neighbors   of $v$.
\end{theorem}

 Theorem \ref{recursiveformula1} shows that the size function  $f({\cal K},{m})$   can be calculated through the term $f({\cal K},{m-1})$   and  the terms  related to  some  subgraphs of $\cal K$. Below, we discuss the explicit formula of   $f({\cal K},{m})$. First, we have the following corollary.

%\begin{corollary}\label{emptycore}
%   Let ${\cal K}_{\emptyset}$ be a null graph. We have $f({{\cal K}_{\emptyset}}^{m+})=m!$.
%\end{corollary}

% Because  the extended graph of ${\cal K}_{\emptyset}$ is a completed graph,  Corollary \ref{emptycore} holds obviously.

 \begin{corollary}\label{solution0}
 Let ${\cal K}$ be an undirected chordal graph. The formula   $f({\cal K},{m})$ defined in Equation (\ref{sizecore}) {is  a  polynomial    divisible by $m!$.}
 \end{corollary}

Consider the recursive formula in Equation (\ref{iterizeformul}), the second term in the right side is crucial to derive  the explicit formula of $f({\cal K},{m})$.
 Define
\begin{equation}\label{iterizeformu2} g({\cal K},m):= \frac{1}{m!} \sum_{v\in V }  f ({\cal K}_{N_v},{m}) \frac{\mbox{Size}({\cal K}^{(v)})}{\mbox{Size} ({\cal K}_{N_v})}.
\end{equation}
If  $\cal K$ is an undirected chordal graph, its induced subgraph ${\cal K}_{N_v}$ is also an undirected chordal graph. According to Corollary \ref{solution0},  the formula $f ({\cal K}_{N_v},{m}) $ is  a  polynomial    divisible  by  $m!$, it follows that   the formula $g({\cal K},m)$ defined in Equation (\ref{iterizeformu2}) is a  polynomial of $m$.
Let  $d$ be the degree of polynomial $g({\cal K},m)$, according to Corollary \ref{solution0}, $g({\cal K},m)$  can be represented by

\begin{equation}\label{incrementg}g({\cal K},m)=\sum_{i=1}^{d+1}\gamma_{i} m^{i-1}.\end{equation}

Given the polynomial $g({\cal K},m)$, the following theorem shows the explicit formula of $f({\cal K},{m})$.

  \begin{theorem}\label{solution1}  Let ${\cal K}$ be an undirected chordal graph, $\{\gamma_i, i=1,2,\cdots,d+1\}$ be the coefficients of the polynomial $g({\cal K},m)$ defined in Equation (\ref{incrementg}), and let $a_{ij}=(-1)^{j-i}{ j \choose {i-1}}$ for   any $i\leq j$. We have, for any $m\ge 0$,
\begin{equation}\label{solution1f}{f({\cal K},{m})}=\left(\beta_0+\sum_{i=1}^{d+1}\beta_{i} m^{i}\right){m!}, \end{equation}
where  $\beta_0=  \mbox{Size}({\cal K})$, $\beta_{d+1}= { \gamma_{d+1} }/{a_{d+1,d+1}}$, and  $\beta_{i} =  ({  { \gamma_{i}-\sum_{j=i+1}^{d+1}a_{i,j}\beta_{j} } })/{{a_{i,i}}},$   for any integer $i \in [1,d] $.
\end{theorem}

According to  Theorem \ref{solution1}, to obtain  the explicit formula of   $f({\cal K},m)$ for an undirected chordal graph $\cal K$, we just need to calculate the size  $\mbox{Size}(\cal K)$, and the polynomial $g({\cal K},m)$ defined in Equation (\ref{incrementg}). The algorithms for general core graphs $\cal K$ will be introduced in   Section \ref{algorithmsec}.
Below, we discuss the formulas for some specific types of    undirected chordal graphs.

 When   an  undirected chordal graph  contains some isolated vertices,  these vertices can be removed and  the corresponding  size formula  can be obtained as follows.

\begin{corollary}[Isolated vertices]\label{isolated}   The   graph $\cal K$ is composed of   an undirected chordal graph ${\cal K}_1$  and $j$ isolated vertices. We have

\begin{equation}\label{isonodes}f({\cal K},{m})= f({\cal K}_1,{m}) + j\cdot \mbox{Size}({\cal K}_1)\cdot m m!.\end{equation}
Especially, when ${\cal K}_1$ is a null graph,    we have $f({\cal K},{m})=(jm+1)m!$.
\end{corollary}

A tree is a connected graph without cycle, and a \emph{tree plus} graph is generated by adding one more edge to a tree. We give four explicit size formulas for four specific types of   undirected chordal graphs  in Corollary \ref{threetype}.

\begin{corollary}\label{threetype} Let $\cal K$ be an undirected chordal with $p$ vertices.
\begin{enumerate}
  \item If   ${\cal K}$ is a null graph, we have $f({{\cal K}},{m})=m!$.
  \item   If $\cal K$ is a tree, we have $f({\cal K},{m})= [(p-1)m^2+(2p-1)m+p]m!$.
  \item   If $\cal K$ is a tree plus, we have $f({\cal K},{m})= [ m^3+2pm^2+(4p-1)m+2p]m!$.
  \item   If  $\cal K$ is composed of isolated edges, we have $f({\cal K},{m})= 2^{p/2-1}(pm^2/2+3pm/2+2)m!$.
\end{enumerate}
\end{corollary}

\begin{center}
\begin{table}[h]
\begin{center}
\resizebox{\textwidth}{!}{
\begin{tabular}{cccr||cccr}
  \hline
  id&$(n',p)$& $\cal K$ & $f({\cal K},{m})/m!$&id&$(n',p)$& $\cal K$ & $f({\cal K},{m})/m!$\\\hline
  1& (1, 2)& \unitlength=0.4mm \begin{tabular}{c} \begin{picture}(25,20)(0,0) \thicklines
\put(0,10){\circle*{3}} \put(20,10){\circle*{3}}
 \end{picture} \end{tabular} &$ 2m + 1 $ &
9&(4,5)&\unitlength=0.4mm \begin{tabular}{c} \begin{picture}(25,20)(0,0) \thicklines
\put(0,15){\circle*{3}} \put(0,0){\circle*{3}} \put(20,0){\circle*{3}}
\put(10,15){\circle*{3}} \put(30,15){\circle*{3}}
 \put(10,15){\line(1,0){20}}
\put(0,0){\line(1,0){20}} \put(0,0){\line(2,1){30}}
\put(0,0){\line(2,3){10}} \put(20,0){\line(2,3){10}}
\put(20,0){\line(-2,3){10}}
\end{picture} \end{tabular} & $24m  + (m + 4)\cdots(m+1)
$

 \\\hline

 2&(2,3)&\unitlength=0.4mm \begin{tabular}{c} \begin{picture}(25,20)(0,0) \thicklines
\put(0,0){\circle*{3}} \put(20,0){\circle*{3}}
\put(10,15){\circle*{3}}
\put(0,0){\line(1,0){20}}
\end{picture} \end{tabular}&$m^2 + 5m + 2 $&

10&(5,4)&\unitlength=0.4mm \begin{tabular}{c} \begin{picture}(25,20)(0,0) \thicklines
\put(0,0){\circle*{3}} \put(20,0){\circle*{3}}
\put(10,15){\circle*{3}} \put(30,15){\circle*{3}}
\put(0,0){\line(2,3){10}}
\end{picture} \end{tabular} & $  m^2 + 7m + 2
$
\\\hline

3&(3,3)&\unitlength=0.4mm \begin{tabular}{c} \begin{picture}(25,20)(0,0) \thicklines
\put(0,0){\circle*{3}} \put(20,0){\circle*{3}}
\put(10,15){\circle*{3}}
\end{picture} \end{tabular} & $3m + 1$ &

11&(5,5)&\unitlength=0.4mm \begin{tabular}{c} \begin{picture}(25,20)(0,0) \thicklines
\put(0,15){\circle*{3}} \put(0,0){\circle*{3}} \put(20,0){\circle*{3}}
\put(10,15){\circle*{3}} \put(30,15){\circle*{3}}
 \put(10,15){\line(1,0){20}}
\put(0,0){\line(1,0){20}} \put(0,0){\line(2,1){30}}
\put(0,0){\line(2,3){10}} \put(20,0){\line(2,3){10}}
\end{picture} \end{tabular} & $ 2m^3 + 11m^2 + 29m + 10 $
\\\hline

4&(3,4)&\unitlength=0.4mm \begin{tabular}{c} \begin{picture}(25,20)(0,0) \thicklines
\put(0,0){\circle*{3}} \put(20,0){\circle*{3}}
\put(10,15){\circle*{3}} \put(30,15){\circle*{3}}
 \put(10,15){\line(1,0){20}}
\put(0,0){\line(1,0){20}} \put(0,0){\line(2,1){30}}
\end{picture} \end{tabular} &  $ 3m^2 +7m+4$&
12&(5,5)&\unitlength=0.4mm \begin{tabular}{c} \begin{picture}(25,20)(0,0) \thicklines
\put(0,15){\circle*{3}} \put(0,0){\circle*{3}} \put(20,0){\circle*{3}}
\put(10,15){\circle*{3}} \put(30,15){\circle*{3}}
 \put(10,15){\line(1,0){20}}
\put(0,0){\line(1,0){20}} \put(0,0){\line(2,1){30}}
\put(0,0){\line(0,1){15}} \put(20,0){\line(2,3){10}}
\end{picture} \end{tabular} &$  m^3 + 10m^2 + 19m + 10 $
\\\hline

5&(3,4)&\unitlength=0.3mm
   \begin{tabular}{c} \begin{picture}(25,35)(0,0) \thicklines
\put(0,0){\circle*{3}} \put(20,0){\circle*{3}}
\put(10,15){\circle*{3}} \put(10,30){\circle*{3}}
\put(0,0){\line(1,0){20}} \put(0,0){\line(1,3){10}} \put(20,0){\line(-1,3){10}}
\end{picture} \end{tabular}& $  m^3 + 6m^2 + 17m + 6 $&
13&(5,5)&\unitlength=0.4mm \begin{tabular}{c} \begin{picture}(25,20)(0,0) \thicklines
\put(0,15){\circle*{3}} \put(0,0){\circle*{3}} \put(20,0){\circle*{3}}
\put(10,15){\circle*{3}} \put(30,15){\circle*{3}}
 \put(10,15){\line(1,0){20}}
\put(0,0){\line(1,0){20}} \put(0,0){\line(2,1){30}}
  \put(20,0){\line(2,3){10}} \put(10,15){\line(-1,0){10}}
\end{picture} \end{tabular} & $ m^3 + 10m^2 + 19m + 10 $
\\\hline

6&(4,4)&\unitlength=0.4mm \begin{tabular}{c} \begin{picture}(25,20)(0,0) \thicklines
\put(0,0){\circle*{3}} \put(20,0){\circle*{3}}
\put(10,15){\circle*{3}} \put(30,15){\circle*{3}}
\put(0,0){\line(2,3){10}} \put(20,0){\line(2,3){10}}
\end{picture} \end{tabular} & $ 4m^2 + 12m + 4 $ &
14&(5,6)&\unitlength=0.4mm \begin{tabular}{c} \begin{picture}(30,25)(0,0) \thicklines
\put(0,0){\circle*{3}} \put(0,20){\circle*{3}} \put(10,0){\circle*{3}}
\put(10,20){\circle*{3}} \put(20,10){\circle*{3}} \put(30,10){\circle*{3}}

  \put(0,0){\line(0,1){20}}
   \put(0,0){\line(1,0){10}}
  \put(0,20){\line(1,-2){10}}
  \put(0,20){\line(2,-1){20}}
  \put(0,20){\line(1,0){10}}
  \put(10,0){\line(1,1){10}}
  \put(10,0){\line(0,1){20}}
   \put(10,20){\line(1,-1){10}}
   \put(30,10){\line(-2,1){20}}
   \put(30,10){\line(-1,0){10}}

\end{picture} \end{tabular} &$ m^4 + 14m^3 + 55m^2 + 82m + 40 $
\\\hline

7&(4,4)&\unitlength=0.4mm \begin{tabular}{c} \begin{picture}(25,20)(0,0) \thicklines
\put(0,0){\circle*{3}} \put(20,0){\circle*{3}}
\put(10,15){\circle*{3}} \put(30,15){\circle*{3}}

\put(0,0){\line(1,0){20}} \put(0,0){\line(2,3){10}}
\end{picture} \end{tabular} & $ 2m^2 + 8m + 3 $&
15&(5,6)&\unitlength=0.4mm \begin{tabular}{c} \begin{picture}(30,25)(0,0) \thicklines
\put(0,0){\circle*{3}} \put(0,20){\circle*{3}} \put(10,0){\circle*{3}}
\put(10,20){\circle*{3}} \put(20,10){\circle*{3}} \put(30,10){\circle*{3}}

  \put(0,0){\line(0,1){20}}
  \put(0,20){\line(1,-2){10}}
  \put(0,20){\line(2,-1){20}}
  \put(0,20){\line(1,0){10}}
  \put(10,0){\line(1,1){10}}
  \put(10,0){\line(0,1){20}}
   \put(10,20){\line(1,-1){10}}
   \put(30,10){\line(-2,1){20}}
   \put(30,10){\line(-1,0){10}}
   \put(30,10){\line(-2,-1){20}}
\end{picture} \end{tabular} &$(m + 1)(2m + 3)(m^2 + 7m + 16)
   $

\\\hline

8&(4,5)&\unitlength=0.4mm \begin{tabular}{c} \begin{picture}(25,20)(0,0) \thicklines
\put(0,15){\circle*{3}} \put(0,0){\circle*{3}} \put(20,0){\circle*{3}}
\put(10,15){\circle*{3}} \put(30,15){\circle*{3}}
 \put(10,15){\line(1,0){20}}
\put(0,0){\line(1,0){20}} \put(0,0){\line(2,1){30}}
\put(0,0){\line(2,3){10}} \put(20,0){\line(2,3){10}}
\put(10,15){\line(-1,0){10}}
\end{picture} \end{tabular} & {  $(m + 1)(m + 4)(2m + 3)
$}&

16&(5,6)&\unitlength=0.4mm \begin{tabular}{c} \begin{picture}(30,25)(0,0) \thicklines
\put(0,0){\circle*{3}} \put(0,20){\circle*{3}} \put(10,0){\circle*{3}}
\put(10,20){\circle*{3}} \put(20,10){\circle*{3}} \put(30,10){\circle*{3}}

\put(0,0){\line(1,0){10}}
\put(0,0){\line(1,2){10}}
  \put(0,0){\line(2,1){20}}
  \put(0,0){\line(0,1){20}}
  \put(0,20){\line(1,-2){10}}
  \put(0,20){\line(2,-1){20}}
  \put(0,20){\line(1,0){10}}
  \put(10,0){\line(1,1){10}}
  \put(10,0){\line(0,1){20}}
   \put(10,20){\line(1,-1){10}}
\end{picture} \end{tabular} & $120m  +  (m + 5)\cdots(m+1)$
\\\hline
\end{tabular}}
\caption{The explicit formulas for all core graphs with up to five missing edges, $n',p$ are the number of missing edges and the number of vertices in the core graph $\cal K$, respectively.}
\label{exampleformul}
\end{center}
\end{table}

\end{center}

By  Corollary \ref{isolated}, corollary \ref{threetype} and Theorem \ref{solution1f}, we can obtain the size formula $f({\cal K},m)$ given an undirected chordal graph $\cal K$. \citet{he2015counting} give  two explicit size formulas for essential graphs with one or two missing edges; here we do the same for core graphs with at most five missing edges. In Table \ref{exampleformul}, we list  all core graphs with up to five missing edges, together with their corresponding size formulas.
 %The number of missing edges ($n'$), the number of vertices ($p$), core graph ($\cal K$) and the corresponding size formula  $f({\cal K},{m})/m!$  are displayed.
We give an example to demonstrate the derivation of these formulas. Consider  the last (with id 16) core graph in Table \ref{exampleformul}, $\cal K$  is composed of a completed graph with five vertices (${\cal K}_1$) and one isolated vertex. We have Size$({\cal K}_1)=120$ and $f({\cal K}_1,m)=(m+5)!$ from Lemma \ref{domk},   it follows  $f({\cal K},m)/m!=[(m+5)!+120mm!]/m!= 120m  +  (m + 5)\cdots(m+1)$ by Corollary \ref{isolated}.

Given an undirected connected chordal graph $\cal U$, when its core graph $\cal K$ is small, we can calculate  $g({\cal K},m)$  directly following its definition in Equation (\ref{iterizeformu2}), and then obtain   the explicit formula of $f({\cal K},{m})$ according to Theorem \ref{solution1f}. However, when the core graph is large,  the derivation of  $g({\cal K},m)$   becomes more complicated. In the next section, we will provide an algorithm  to derive  the explicit formulas of $f({\cal K},{m})$ for a general core  graph $\cal K$.

\subsection{Algorithms}\label{algorithmsec}

In this section, we  introduce two main algorithms. The algorithm  {SizeF}$({{\cal K}})$ in Algorithm \ref{count2}  gives  the  explicit formula  of $ f({{\cal K}},{m})$ for       an undirected chordal graph $\cal K$.  The algorithm Size$({\cal C})$ in Algorithm \ref{count} counts  the size of the Markov equivalence class represented by an essential graph   $\cal C$.  Both Algorithm \ref{count2} and Algorithm \ref{count} call each other recursively.
%The sub-algorithms used in these two algorithms, including Algorithm \ref{postroot},  Algorithm \ref{count21}, Algorithm\ref{count22} and Algorithm \ref{count23} also be  presented in   this section.

\begin{algorithm}[h]
\caption{{SizeF}$({{\cal K}})$}
\label{count2}

\KwIn{ ${\cal K}$, an undirected chordal graph;}
\KwOut{ $ f({{\cal K}},{m})$, a polynomial of $m$.}

Let \emph{type} be the type of ${\cal K}$ and $p$ be the number of vertices in ${\cal K}$\;
\Switch{type}{\lCase{null graph}{\Return{$ m!$}}\;
  \lCase{tree}{\Return{$[(p-1) m^2+(2n-1)m+p]m!$}}\;
  \lCase{tree-plus}{\Return{$( m^3+2pm^2 +(4p-1)m+2p)m!$}}\;
  \lCase{isolated-edge graph}{\Return{$ 2^{p/2-1}(pm^2/2+3pm/2+2)m!$}}
   }

Let $w$  be the number of dominating vertices in ${\cal K}$; remove these vertices from $\cal K$\;

\If {$w>0$}{

${h}(m) \leftarrow$  {SizeF}$({\cal K})$\;
\Return{h(m+w)}
}

Let $k$ be the number of isolated vertices  in ${\cal K}$;
remove these vertices from ${\cal K}$\;

\If {$k>0$} {
\Return{${\emph{SizeF}}({{\cal K}})+ \emph{Size}({\cal K})kmm!$}, (see  {Size}(${\cal K}$) in  Algorithm \ref{count})\;
}

%\If {  $H$ is empty, a tree or  a tree plus}
%{\Return{SimpleCase($H$)}( see Algorithm \ref{count21} in Appendix \ref{algorithmadd})}

\Return{SizeGF(${\cal K}$)}, (see SizeGF(${\cal K}$) in  Algorithm \ref{count22})\;

\end{algorithm}
\begin{algorithm}[h]
\caption{{{Size}}$({\cal C})$}
\label{count}

\KwIn{ $\cal C$, an essential graph;}
\KwOut{ the size of Markov equivalence classes represented by   $ {\cal C}$.}
Let ${\cal C}_1,\cdots,{\cal C}_J$
be all of chain components of $\cal U$;  for any integer $0\leq J\leq J$,  $m_j$ is the number of dominating vertices in ${\cal C}_j$ and ${\cal K}_j$ the core graph of ${\cal C}_j$\;
\For{$j \leftarrow 1$ \KwTo $J$}{
  $f_j(m) \leftarrow$  {SizeF}(${\cal K}_i$) \;}
 \Return{$\prod_{j=1}^J f_j(m_j)$.}
\end{algorithm}

 In Algorithm \ref{count2},  we first give the explicit   formula of $f({\cal K},m)$  when $\cal K$ is null, tree, tree-plus or isolated-edge  graph  according to     Proportion \ref{threetype}. Otherwise, when the undirected chordal graph $\cal K$ contains   dominating vertices or isolated vertices, we  simplify the formula derivation  according to Lemma \ref{domk} or Corollary \ref{isolated}, respectively.   Finally, for a general undirected chordal graph $\cal K$, we derive  the explicit  formula of $f({\cal K},{m})$ by  the algorithm called SizeGF(${\cal K}$) in Algorithm \ref{count22}.

The algorithm SizeGF(${\cal K}$) in Algorithm \ref{count22} first calculates the  polynomial $g({\cal K},m)$ defined in Equation (\ref{incrementg}) and then derives  the explicit polynomial $f({\cal K},m)$ according to Theorem \ref{solution1}. Suppose that the undirected chordal graph  $\cal K$   contains $J$ isolated connected subgraphs, we  calculate the  polynomial $g({\cal K},m)$ in the first part of Algorithm \ref{count22} (line 1 to 4) according to Corollary \ref{countgm} as follows.

\begin{corollary}\label{countgm}
  Let $\cal K$ be an undirected chordal graph with $ J $ isolated connected subgraphs, denoted by ${\cal K}_1,\cdots,{\cal K}_J$ respectively, $V({\cal K}_j)$ be the set of vertices in ${\cal }K_j$, and $g({\cal K},m)$ is the polynomial defined in Equation (\ref{incrementg}). We have
\begin{equation}\label{incrementg2}g({\cal K},m)=\sum_{j=1}^J \frac{\mbox{Size}({\cal K})}{\mbox{Size}({\cal K}_j)}\sum_{v\in V({\cal K}_j)}\frac{ f({\cal K}_{j,N_v},{m})}{m!}\frac{ \mbox{Size}({\cal K}_{j}^{(v)})}{\mbox{Size}({\cal K}_{j,N_v})},
\end{equation}
where  ${\cal K}_{j}^{(v)}$ is the $v$-rooted essential graph of ${\cal K}_j$, and ${\cal K}_{j,N_v}$ is the induced subgraph of ${\cal K}_{j}$ on the neighbours of $v$.
\end{corollary}

\begin{algorithm}[h]
\SetAlgoRefName{1.1}
\caption{{{SizeGF}}$({{\cal K}})$}
\label{count22}

\KwIn{ ${\cal K}$,  an undirected chordal graph;}
\KwOut{ $ f({{\cal K}},{m})$, a polynomial of $m$.}

Let ${{\cal K}}_1,\cdots,{{\cal K}}_J$ be  $J$ UCCGs in ${\cal K}$, $V({\cal K}_j)$ be the vertex set of  ${\cal }K_j$\;
Set $S_{{\cal K}_{j}}^{(v)}\leftarrow {\mbox{Size}({\cal K}_{j}^{(v)})}$ for any integer $j\in [1,J]$ and any $v \in V({{\cal K}}_j)$\;
$S_{{\cal K}_j}\leftarrow \sum_{v\in V({\cal K}_j)}  {\mbox{Size}({\cal K}_{j}^{(v)})}$, $S_{\cal K}\leftarrow  \prod_{j=1}^J S_{{\cal K}_j}$\;

$g(m)\leftarrow \sum_{j=1}^J \frac{ S_{\cal K}}{S_{{\cal K}_j}}\sum_{v\in V({\cal K}_j)}\frac{\mbox{SizeF}({\cal K}_{j,N_v} )}{m!}\frac{ S_{{\cal K}_{j}}^{(v)}}{\mbox{Size}({\cal K}_{j,N_v})}$ and
denote it as $\sum_{i=1}^{d+1}\gamma_{i} m^{i-1}$\;
Set
$\beta_0\leftarrow S_{\cal K}$;  $\beta_{d+1}\leftarrow { \gamma_{d+1} }/{a_{d+1,d+1}} $ and $a_{ij}\leftarrow (-1)^{j-i}{j \choose {i-1}}$ for $i\leq j\leq d+1$\;
\For {$i \leftarrow d$ \KwTo $1$} {$\beta_{i} \leftarrow \frac{ \gamma_{i}-\sum_{j=i+1}^{d+1}a_{i,j}\beta_{j} }{a_{i,i}}$\;}
\Return{$\sum_{i=0}^{d+1}\beta_i m^i$m!.}
%\Return{$g(m)$}
\end{algorithm}

In  Algorithm \ref{count22}, we need to calculate   $\mbox{Size}({\cal K}_{j}^{(v)})$ for some $j$ and $v$, which are the sizes of Markov equivalence classes represented by   rooted essential graphs. \citet{he2015counting} propose  an algorithm
called ChainCom  to construct the rooted essential graph and  all of its chain components for a UCCG   and a root vertex.  We give ChainCom  in Algorithm \ref{postroot} in Appendix  for the completion of the paper.

In Algorithm \ref{count}, we first find the core graphs of the chain components of the essential graph $\cal C$, then calculate the size of the corresponding Markov equivalence class  by using the formulas obtained from  Algorithm  \ref{count2}. When  some  subgraphs of these chain components contain dominating vertices, formula-based  size counting   will display its advantages; this will be studied experimentally in the next section.

\section{Experimental Results}\label{chap4}

 In this section, we introduce the implementation of  the formula derivation and formula-based counting algorithms, and conduct experiments to evaluate  the  formula-based size counting algorithm  proposed in Section \ref{chap3}.  All experiments are  run  on a linux server  at Intel 2.0GHz. These experiments display  that the proposed alorithms greatly speed up   the size counting, especially   when the corresponding UCCGs  contain  dense subgraphs.

\subsection{A Python package for size formula derivation}

We developed a  Python package named countMEC to derive  the size formulas  and to count the sizes of Markov equivalence classes  based on these formulas. The {symbolic computation} in  countMEC depends on the python package  \emph{sympy}. {The following   example demonstrates  the  usage of the package countMEC.}

\begin{verbatim}
1. from countMEC import *
2. G=ran_conn_chordal_graph(15,95)
3. K=core_graph(G)
4. F=SizeF([K])
5. S=Size(G)
\end{verbatim}

\begin{figure}[h]
  \centering
  \begin{minipage}[b]{0.45\linewidth}
  \includegraphics[width=\textwidth]{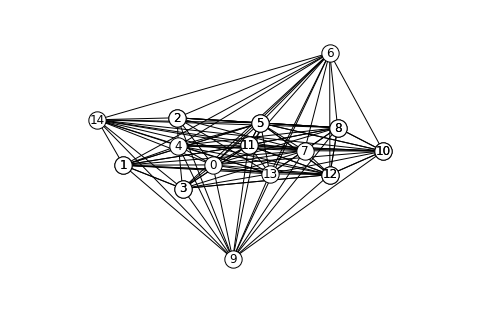}
  \vspace{-1cm}
  $$G$$
  \end{minipage}
  \begin{minipage}[b]{0.45\linewidth}
   \includegraphics[width=\textwidth]{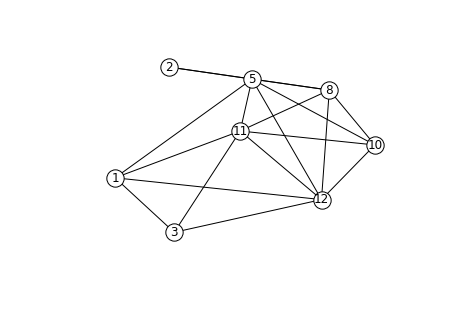}
    \vspace{-1cm}
  $$K$$
  \end{minipage}
  \caption{A UCCG $G$ with 15 vertices and 95 edges and its   core graph  $K$.} \label{ggg}
  \end{figure}

 In this example,  we first import the package countMEC, and  randomly generate  a UCCG $G$ with 15 vertices and 95 edges. The graph $G$ is shown in  the left of Figure \ref{ggg}. Then, we get the core graph of $G$, denoted by $K$, which is shown  in the right of  Figure \ref{ggg}. The  graph $G$ contains $7$ dominating vertices and the  core graph $K$  just contains 8 vertices and 17 edges. In the fourth line, we call  SizeF($\cdot$) (Algorithm \ref{count2}); it outputs the following size formula: $F(m)=\left(m^{3} + 16 m^{2} + 77 m + 108\right) (m+2)!$. In the last line, we call Size($\cdot$) (Algorithm \ref{count}) and  get  $S=643749120$, which is the size of  $G$. It's easy to check that $S=F(7)$. In this example, it takes   0.5 second to count size using the proposed formula-based algorithm, while  440 seconds are taken with the method introduced in \citet{he2015counting}; we will compare the time complexities of two methods thoroughly in the next section.

\subsection{Formula-based  size counting}\label{ch4.2}

In this section, we  experimentally compare the time complexity of our proposed counting algorithms to the benchmark algorithm introduced in \citet{he2015counting}.
Let ${\mathbb {U}}_p^{n}$ be the set of Markov equivalence classes with $p$ vertices and $n$ edges.  We obtain   random choral graphs from ${\mathbb {U}}_p^{n}$ following \citet{he2015counting}.  First, we construct a  tree by  connecting two vertices  (one is sampled  from the connected vertices and the other from the isolated vertices) sequentially until all $p$ vertices are connected. Then, we randomly insert an  edge  such that the resulting graph is chordal, repeatedly until the number of edges reaches $n$.  Repeating this procedure $N$ times, we obtain $N$  samples from ${\mathbb {U}}_p^{j}$ for each integer $j(\leq n)$.

\begin{figure}[h]
  \centering
  \includegraphics[scale=0.55]{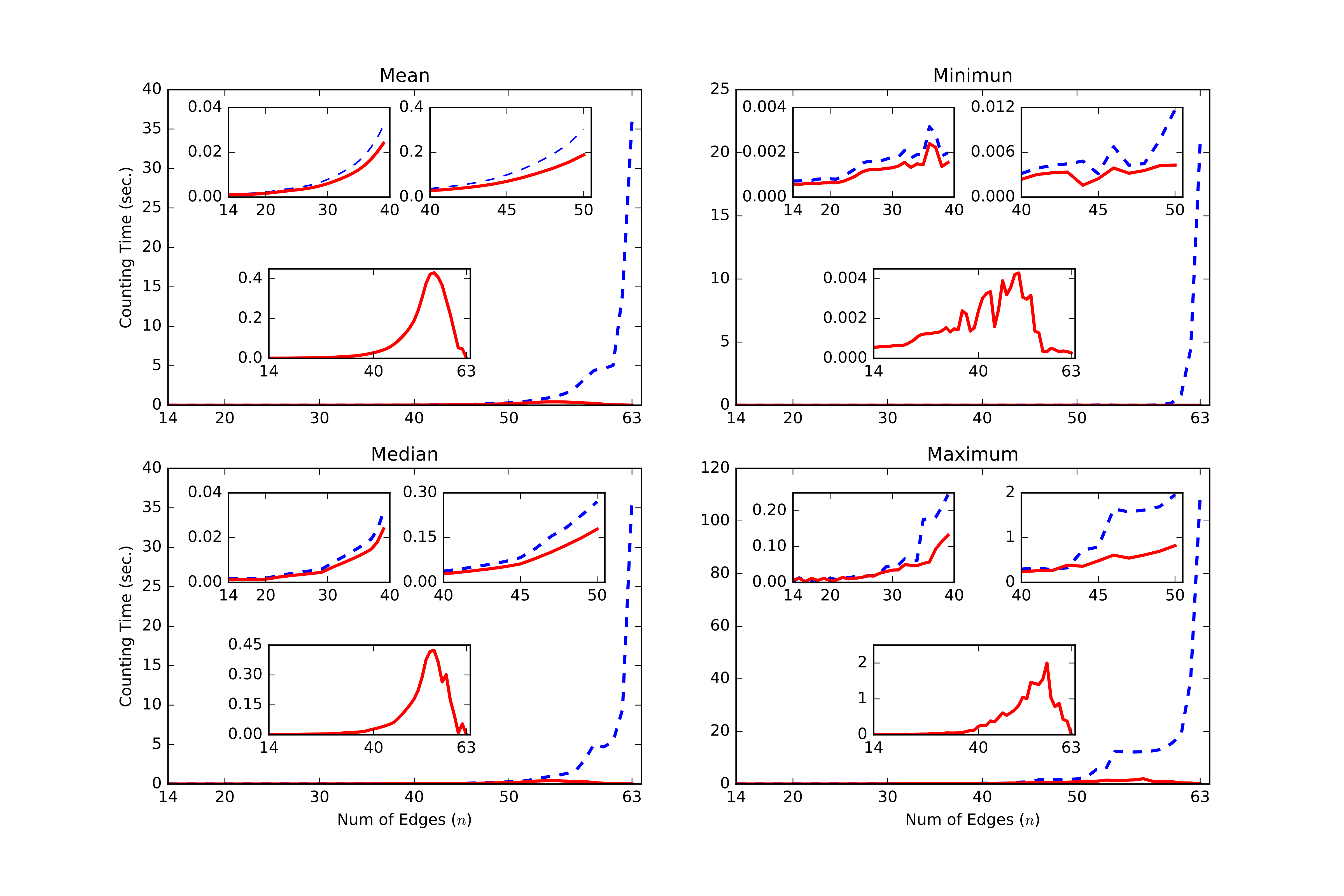}

  \caption{The mean, the minimum, the median and the maximum of counting time of Markov equivalence classes with 12 vertices and $n$ edges.}\label{edgetime}
\end{figure}

We first consider the UCCGs in ${\mathbb {U}}_p^{n}$ with $p\leq 12$ for each integer $n\in [p+2,p(p-1)/2-3]$. Because the results have the similar patterns for different $p$,  we just report the experiments for $p=12$ in this paper.  Based on the $10^5$ samples from ${\mathbb {U}}_{12}^{n}$  for each integer $n\in [14,63]$, we plot the mean, the minimum, the median, and the maximum of  the counting time used by the benchmark algorithm (blue dashed lines)   and   by the proposed  Algorithm \ref{count} (red solid lines)
 in  four panels of Figure \ref{edgetime}, respectively.  In each panel of Figure  \ref{edgetime}, the main window displays all results ($n \in [14,63]$) of both algorithms, the two upper  sub-windows display the results of both algorithms for  $n \in [14,39]$ and $n \in [40,50]$, respectively, and  the lower sub-window  displays the results of   Algorithm \ref{count} again with a proper   size-coordinate.

 We see that the counting time (mean, minimum, median, and maximum) of the benchmark algorithm is increasing in the number of edges ($n$); size counting based on  benchmark algorithm becomes much time-consuming when the graphs are dense. Meanwhile, the time used by Algorithm \ref{count},  increases first, and then decreases with  the number of edges. Figure \ref{edgetime} shows that size counting based on Algorithm \ref{count} keeps   efficient for  both sparse and dense graphs.

 We also study the sets  ${\mathbb U}_{p}^{n}$ that contain UCCGs with tens of vertices under sparsity constraints. The number of vertices  $p$ is set to $20,50$, and $100$, and the number of  edges  $n$ is set to $rp$ where $r$ is the ratio of $n$  to $p$. For each $p$, we consider three ratios: 3, 4 and 5. The    graphs in  ${\mathbb U}_{p}^{rp}$ are sparse since $r\leq 5$. For each pair of $(p,r)$,  $10^5$ UCCGs are generated randomly and then sorted in ascending order according to  the counting time used by benchmark algorithm. The ordered  $10^5$ UCCGs are divided into four subsets. The subset $S_1$ contains the  first 500 UCCGs, $S_2$ contains the next 49500 UCCGs,  $S_3$ contains the next 49500 UCCGs after $S_2$, and $S_4$ contains  the last 500 UCCGs. For each subset, we report the average  of counting time and the average  of their ratios in Table \ref{p20} for the benchmark algorithm ($T_1$) and  the proposed algorithm \ref{count} ($T_2$). We see that on average, (1) the proposed Algorithm  \ref{count} is faster than the benchmark algorithm in all cases, (2) the more edges the UCCGs have ($r$ from $3$ to $5$), or the more  time benchmark algorithm used (subset from $S_1$ to $S_4$),  the  smaller $T_2/T_1$, that is,  the   higher speedup Algorithm \ref{count} achieved. For example, consider the subsets $S_4$ and $r=5$, the average counting time is shorten rapidly for all $p\in\{20, 50,100\}$, the average of ratios  $T_2/T_1$ are also reduced to nearly 0.02.

\begin{table}[h]
\center
\resizebox{\textwidth}{!}{%
\begin{tabular}{c|ccccrrrrrrrrrrr}
\multicolumn{2}{c}{\hspace{-2ex}$p$}&\multicolumn{2}{c}{ \hspace{-4ex} Subset}&\multicolumn{11}{c}{$r$}   \\ \cline{1-1}\cline{3-3}\cline{5-15}
\multicolumn{2}{c}{}&\multicolumn{2}{c}{ }& \multicolumn{3}{c}{$3$} & & \multicolumn{3}{c}{$4$}  & &  \multicolumn{3}{c}{$5$}   \\  \cline{5-7} \cline{9-11} \cline{13-15}
\multicolumn{4}{c}{}& $T_1$ &$T_2$ &$T_2/T_1$& &$T_1$ &$T_2$&$T_2/T_1$&& $T_1$ &$T_2$&$T_2/T_1$    \\

\multirow{5}{*}{20 }   && $S_1$  & &  0.01 & 0.01 & 0.76 & &   0.02 &  0.02 & 0.76 & &     0.02 &   0.02 & 0.75  \\
                      & & $S_2$  & &  0.03 & 0.03 & 0.77 & &   0.17 &  0.13 & 0.74 & &     1.47 &   0.96 & 0.67   \\
                      & & $S_3$   & &  0.10 & 0.07 & 0.74 & &   1.03 &  0.52 & 0.63 & &    21.73 &   4.38 & 0.38   \\
                      & & $S_4$           & &  0.68 & 0.32 & 0.55 & &  21.14 &  2.23 & 0.17 & &   954.22 &  10.92 & 0.02    \\
                    [-2ex]
 \multicolumn{15}{c}{}\\
 \multirow{5}{*}{50}   &&  $S_1$   & &  0.07 & 0.05 & 0.79 & &   0.19 &  0.15 & 0.79 & &     0.74 &   0.56 & 0.76  \\
                     &&    $S_2$  & &  0.18 & 0.14 & 0.77 & &   0.77 &  0.55 & 0.73 & &     5.82 &   3.21 & 0.59   \\
                       &&  $S_3$    & &  0.55 & 0.40 & 0.74 & &   5.18 &  2.39 & 0.59 & &   113.22 &  17.46 & 0.34   \\
                        && $S_4$           & &  5.62 & 2.10 & 0.41 & & 238.98 & 18.80 & 0.15 & & 17598.39 & 128.65 & 0.02    \\
                    [-2ex]
 \multicolumn{15}{c}{}\\
 \multirow{4}{*}{ 100}
                  && $S_1$ & &  0.26 & 0.21 & 0.80 & &   0.73 &  0.58 & 0.80 & &     3.18 &   2.27 & 0.71   \\
                   &&$S_2$   & &  0.78 & 0.60 & 0.77 & &   2.92 &  2.05 & 0.71 & &    21.86 &  10.90 & 0.53  \\
                  && $S_3$  & &  2.25 & 1.63 & 0.74 & &  19.96 &  9.04 & 0.56 & &   429.61 &  55.63 & 0.27    \\
                 &&  $S_4$ & & 21.14 & 7.81 & 0.43 & & 897.18 & 59.59 & 0.10 & & 59093.25 & 516.44 & 0.02     \\

\end{tabular}
}%
\caption{The average of counting time ($T_1$ for benchmark algorithm  and  $T_2$ for Algrithm \ref{count}) and   ratios ($T_2/T_1$) for  UCCGs with $p$ vertices and   $pr$ edges in different subsets.}\label{p20}
\end{table}

 We   have to point out that the choral graphs generated in our experiments  might not be uniformly distributed in the space of chordal graphs and that   the  results in Figure \ref{edgetime} and Table \ref{p20} are not accurate estimations of  expectations of the corresponding statistics.

\section{Conclusion and discussion}\label{chap5}

In this paper, we propose a method  to derive  the size formulas  of Markov equivalence classes  and to count the sizes based on these formulas.  A core graph of an undirected connected chordal graph is introduced and the size formula derivation based on the core graph is proposed. We discuss  both recursive  and  explicit forms of the size formulas and give algorithm to derive  these formulas. Comparing to the benchmark counting algorithm, the proposed algorithm can generate more  size formulas efficiently,  and by these formulas,  size counting  is accelerated dramatically  when  the essential graph   contains non-sparse undirected  subgraphs.

%RSolve[{a[n+1]-(n-3) a[n]==6 (n-3)!+3 (n-1)!,a[4]=6},a[n],n]
%RSolve[{a[n+1]-(n-3) a[n]==6(n-2)!-2 (n-3)!,a[4]=4},a[n],n]
%RSolve[{a[n+1]-(n-2) a[n]==3 (n-2)!,a[4]=4},a[n],n]

\acks{  This work was supported partially by  NSFC  (11671020, 11101008,  71271211). %We are very grateful to Bin Yu for her helpful suggestions and comments about this work.
%DPHEC-20110001120113, US NSF grants DMS-1107000, CDS\&E-MSS 1228246, ARO grant W911NF-11-1-0114, AFOSR Grant FA 9550-14-0016, and the Center for Science of Information (CSoI, a US NSF Science and Technology Center) under grant agreement CCF-0939370.
 }

 \appendix
\section{Algorithm  ChainCom(${\cal U},v$)}

For the completion of the paper,   we give the algorithm   ChainCom$({\cal U}, v)$ in Algorithm \ref{postroot}, which is introduced in    \citet{he2015counting}, to   construct the rooted essential graph ${\cal U}^{(v)}$ and  all of its chain components.

\begin{algorithm}[h]
%\SetAlgoRefName{2.1}
\caption{\emph{ChainCom}$({\cal U},v)$}
\label{postroot}

\KwIn{ $\cal U$, a UCCG; $v$, a vertex of $\cal U$.}
\KwOut{  $v-$rooted essential graph of $\cal U$ and  all of its chain components.}

Set $A=\{v\}$, $B=\tau \setminus v$, ${\mathcal G}={\cal U}$ and ${\cal O}=\emptyset$

\While {$B$ is not empty}{
Set ${T}=\{w: w \mbox{ in } {B} \mbox{ and    adjacent to   $A$} \}$ \;

Orient all edges between $A$ and $  T$ as $c\to t$ in $\cal G$, where $c\in A , t\in   T$\;

%Consider each  undirected edge $y-z$ with $y,z\in T$ repeatedly, orient  $y-z$ to $y\to z$ in $\mathcal G$  if induced subgraph $x \to y - z$  appears in $\mathcal G$ until no more undirected edges can be oriented\;

\Repeat{ no more undirected edges  in ${\cal G}_{T}$ can be oriented
}{
\For {each edge $y-z$ in  the vertex-induced subgraph ${\cal G}_{  T}$}
 {
 \If {$x \to y - z$ in  $\mathcal G$ and $x $ and $z $ are not adjacent in
      $\mathcal G$} {Orient $y-z$ to $y\to z$ in $\mathcal G$ }}

}

Set ${A}=  T$ and ${B}={B}\setminus {  T}$\;
Append all isolated undirected graphs in ${\cal G}_{  T}$ to $\cal O $\;

}

  \Return{ $\cal G$ and ${\cal O}$}
\end{algorithm}
\section{Proofs of Results}
In this section, we provide the proofs of the main results of our paper.

\vspace{0.5cm}
\textbf{{Proof of Proposition  \ref{corepro}}}

Let $v_{i_1}-v_{j_1}$  and    $v_{i_2}-v_{j_2}$ be  two  edges in ${\cal K}^c$. If neither they  share a common vertex, nor they are connected by  an edge,  we have that  $v_{i_1}, v_{j_1}, v_{i_2}, v_{j_2}$ are four distinct vertices and there is no edge between $v_{i_1},v_{j_1}$ and $v_{i_2}, v_{j_2}$. Since that $\bar{\cal K}$ is the complement of $\cal K$, we have  that the four edges, $v_{i_1}-v_{i_2}$, $v_{i_2}-v_{j_1}$, $v_{j_1}-v_{j_2}$, and $v_{j_2}- v_{i_1}$ appear in $\cal K$, and meanwhile, the two edges  $v_{i_1}-v_{j_1}$  and    $v_{i_2}-v_{j_2}$ do not occur in $\cal K$. This implies that no chord exists in the cycle $v_{i_1}-v_{i_2}-v_{j_1}-v_{j_2}-v_{i_1}$ in $\cal K$. It is a contradiction because $\cal K$ is a chordal graph.

Since   no dominating vertices  appear in $\cal K$, for any vertex $v$ in $\cal K$, there exists another vertex in $\cal K$ such that it is not adjacent to $v$. Consequently, there is no isolated vertex in ${\cal K}^c$. Following the proof in the last paragraph, there are  no two edges that occur separatively in two isolated subgraphs of $\cal K$. As a result, ${\cal K}^c$ is a connected graph.
\hfill$\blacksquare$

Before proving Theorem \ref{recursiveformula1}, we give  the following lemma.

\begin{lemma}\label{rooted1}
Let $\cal U$ be an undirected chordal graph over $V$ and  ${\cal U}^{(v)}$  be the   v-rooted graph of $\cal U$. We have that the subgraph of ${\cal U}^{(v)}$ on the neighbors of $v$, denoted by ${\cal U}_{N_v}^{(v)}$, is undirected.
\end{lemma}

\begin{proof}
We can get ${\cal U}^{(v)}$ using  Algorithm \ref{postroot}. Consider any edge, denoted by $v_i- v_j$,  in  ${\cal U}_{N_v}$, $v,v_i$ and $v_j$ form a triangle.   According to  Algorithm \ref{postroot}, $v_i- v_j$ can not be oriented to a directed edge since $v\to v_i-v_j$ is not a induced subgraph of $\cal U$. Therefore, we have that ${\cal U}_{N_v}^{(v)}$  is undirected.
\end{proof}

\vspace{0.2cm}
 \textbf{{Proof of Theorem  \ref{recursiveformula1}}}

Denote the vertices of  ${\cal K}$  as  $V=\{v_1,\cdots,v_p\}$, and  the $m$ extended vertices in ${\cal K}^{m+}$ as $V^\prime=\{v_{p+1},\cdots,v_{p+m}\}$.  From Lemma \ref{calcu151}, we have

\begin{equation}\label{formulp1}
  f\left({{\cal K},{m}}\right)= \sum_{v\in V}  \mbox{Size}\left(({{\cal K}^{m+}})^{(v)}\right)+\sum_{v\in V^\prime}  \mbox{Size}\left(({{\cal K}^{m+}})^{(v)}\right).
\end{equation}
For any $v\in V^\prime$, since $v$ is adjacent to all other vertices in ${{\cal K}^{m+}}$,   from Lemma \ref{rooted1}, we have  $\mbox{Size}\left(({{\cal K}^{m+}})^{(v)}\right)=f\left({{\cal K},{m-1}}\right)$ and
\begin{equation}\label{formulp2}
\sum_{v\in V^\prime}  \mbox{Size}\left(({{\cal K}^{m+}})^{(v)}\right)=m\cdot f\left({{\cal K},{m-1}}\right).
\end{equation}

\begin{figure}
\centering
 \includegraphics[width=0.5\textwidth]{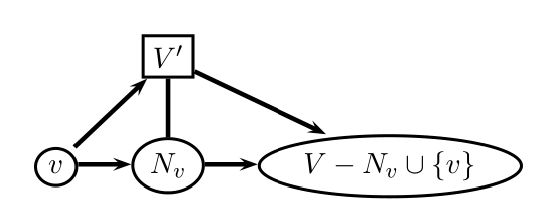} 
\caption{The directions of edges among $v$, $N_v$ and $V^\prime$ and the other vertices in $({{\cal K}^{m+}})^{(v)}$, where $v\to N_v$ represents that each  edge  between $v$  and $N_v$ is directed from $v$  to the vertex in $N_v$, and $V^\prime - N_v$ represents that all edges between $V^\prime$ and $N_v$ are undirected.}\label{figplus1}
\end{figure}

For any  $v\in V$, the neighbor set of $v$ in ${\cal K}^{(m+)}$ is $N_v\cup V^\prime$, from Lemma \ref{rooted1}, ${({\cal K}_{N_v})}^{m+}$ is a chain component of $({{\cal K}^{m+}})^{(v)}$ when $m>0$. According to Algorithm \ref{postroot} and Lemma \ref{rooted1}, the directions of edges among $v$, $N_v$ and $V^\prime$ and the other vertices in $({{\cal K}^{m+}})^{(v)}$  are displayed in Figure \ref{figplus1}. All edges are directed from $ N_v\cup V^\prime$   to $V-N_v\cup \{v\}$ in $({{\cal K}^{m+}})^{(v)}$.  We have

 $$\mbox{Size}\left(({{\cal K}^{m+}})^{(v)}\right)=  \mbox{Size}\left(\left(({{\cal K}^{m+}})^{(v)}\right)_{N_v\cup V^\prime}\right)\mbox{Size}\left(\left(({{\cal K}^{m+}})^{(v)}\right)_{V-N_v\cup \{v\}}\right) $$

First,  according to Lemma  \ref{rooted1}, we can get that $\left(({{\cal K}^{m+}})^{(v)}\right)_{N_v\cup V^\prime}$ is   the same as ${({\cal K}_{N_v})}^{m+}$, thus, $\mbox{Size}\left(\left(({{\cal K}^{m+}})^{(v)}\right)_{N_v\cup V^\prime} \right)=f\left({{\cal K}_{N_v}},{m} \right) $ holds. Then, consider the undirected edges in $\left(({{\cal K}^{m+}})^{(v)}\right)_{V-N_v\cup \{v\}}$, according to Algorithm \ref{postroot}, because all vertices in $V^\prime$ are parents of vertices in $V-N_v\cup \{v\}$,  we have that  $\left(({{\cal K}^{m+}})^{(v)}\right)_{V-N_v\cup \{v\}}$ has the same chain components as $\left({\cal K}^{(v)}\right)_{V-N_v\cup \{v\}}$. As a result,  $\mbox{Size}\left(\left(({{\cal K}^{m+}})^{(v)}\right)_{V-N_v\cup \{v\}}\right)=\mbox{Size}\left(\left({\cal K}^{(v)}\right)_{V-N_v\cup \{v\}}\right)$. Moreover, according to Equation (\ref{formula0}), we have  $\mbox{Size}\left(\left({ \cal K}^{(v)}\right)_{V-N_v\cup \{v\}}\right)=\frac{\mbox{Size}({\cal K}^{(v)})}{\mbox{Size}({\cal K}_{N_v})}$. Consequently, we have
\begin{equation}\label{formulp3}
 f\left(({{\cal K}^{m+}})^{(v)}\right)
 = f({{\cal K}_{N_v}},{m}) \frac{\mbox{Size}({\cal K}^{(v)})}{\mbox{Size}({\cal K}_{N_v})}.
\end{equation}
Theorem \ref{recursiveformula1} holds directly from Equation (\ref{formulp1}), Equation (\ref{formulp2}) and Equation (\ref{formulp3}).
\hfill$\blacksquare$

\vspace{0.5cm}
 \textbf{{Proof of Corollary \ref{solution0}}}

%Given a graph $\cal G$, we denote the vertex set of $\cal G$ as $V({\cal G})$.
For any undirected chordal graph  $\cal K$,  from Theorem \ref{recursiveformula1}, we have
\begin{equation}\label{pformu1}f\left({\cal K},{m}\right)=  m \cdot f\left({\cal K},{m-1}\right)+\sum_{v\in V} f\left( {\cal K}_{N_v},{m}\right) h({\cal K},v)\end{equation}
where $V$ is the set of vertices in $\cal K$,  $h({\cal K},v)$ is an integer function of  $\cal K$ and $v$. Consider  $f(\cdot,\cdot)$ terms   in the right side of  Equation (\ref{pformu1}), we can   calculate them by using  Equation (\ref{pformu1}) again  as follows.

\begin{equation}\label{pformu2}f\left({\cal K},{m-1}\right)=  (m-1) \cdot f\left({\cal K},{m-2}\right)+\sum_{v\in V } f\left( {\cal K}_{N_v},{m-1}\right) h({\cal K},v) \end{equation}
and
\begin{equation}\label{pformu3}f\left({\cal K}_{N_v},{m}\right)=  m \cdot f\left({{\cal K}_{N_v}},{m-1}\right)+\sum_{v^\prime \in N_v } f\left({{\cal K}}_{N^\prime_{v^\prime}},{m}\right) h({\cal K}_{N_v},v), \end{equation}
where $N^\prime_{v^\prime}$ is the neighbor set of $v^\prime$ in ${\cal K}_{N_v}$. Replacing $f\left({\cal K},m-1\right)$ and $f\left( {\cal K}_{N_v},{m}\right)$ in Equation (\ref{pformu1}) by the corresponding terms in Equation (\ref{pformu2}) and Equation (\ref{pformu3}), we can find that $f\left({\cal K},{m}\right)$ is the sum of   the following three types of terms,
\begin{enumerate}
  \item $m(m-1)f({\cal K},{m-2})$,
  \item  $mf({{\cal K}_{N_v}},{m-1})h({\cal K},v)$, for any $v\in V({\cal K})$, and
  \item  $f ({{\cal K}}_{N^\prime_{v^\prime}},{m})h({\cal K}_{N_v},v)$, where $N^\prime_{v^\prime}$ is the neighbor set of $v^\prime$ in ${\cal K}_{N_v}$, for any $v^\prime, v$ such that $v^\prime \in N_v$.
\end{enumerate}

Notice that for any $v,v^\prime$, we have $N^\prime_{v^\prime}\subset N_v\subset V$, so the graphs in above three types of terms  are smaller than that in Equation (\ref{pformu1}). By this way, using Equation (\ref{pformu1}) repeatedly,  we can calculate $f ({\cal K},{m} )$  by smaller graphs. Finally,  $f ({\cal K},{m} )$ can be calculated only by $f ({\cal K},{0} )$ and $f ( ({\cal K}_{\emptyset} ),{k} )$ for $k\leq m$. As a result,  $f ({\cal K},{m} )$ is the sum of  some polynomials of $m$ and each term of the polynomials contains either   $m!f ({\cal K},{0} )$ or $\frac{m!}{k!}f ( ({\cal K}_{\emptyset} ),{k} )$ for $k\leq m$.
Because ${\cal K}_{\emptyset}$ is null graph, $f ( ({\cal K}_{\emptyset} ),{k} )=k!$, we have that   $f({\cal K},{m})$  is  a  polynomial    divisible by $m!$. \hfill$\blacksquare$

\vspace{0.5cm}
 \textbf{{Proof of Theorem \ref{solution1}}}

 We just need to show that Formula  (\ref{solution1f}) is the  solution of   Equation (\ref{iterizeformul}). First, when $m=0$, we have $f({\cal K},{m})=\beta_0=\mbox{Size}({\cal K})$. Theorem \ref{solution1} holds if the following equation holds,
$$
 \left(\beta_0+\sum_{i=1}^{d+1}\beta_{i} m^{i}\right){m!}= m\left(\beta_0+\sum_{i=1}^{d+1}\beta_{i} (m-1)^{i}\right){(m-1)!}+\sum_{i=1}^{d+1}\gamma_{i} m^{i-1}m!.$$
Equivalently,
 \begin{equation}\label{pformu4}
 \sum_{i=1}^{d+1}\beta_{i} m^{i}-\sum_{i=1}^{d+1}\beta_{i} (m-1)^{i}=\sum_{i=1}^{d+1}\gamma_{i} m^{i-1}
 \end{equation}
 Consider the left side of Equation (\ref{pformu4}),
$$ \begin{array}{rl}\sum_{i=1}^{d+1}\beta_{i}\left[ m^{i}-(m-1)^{i}\right]= &\sum_{i=1}^{d+1}\beta_i\left[\sum_{j=0}^{i-1}(-1)^{i-(j+1)}{i \choose j}m^j\right] \\
=&\sum_{j=0}^{d}\sum_{i=j+1}^{d+1}\left[(-1)^{i-(j+1)}{i \choose j}\beta_i m^j\right] \\
=&\sum_{k=1}^{d+1}\left[\sum_{i=k}^{d+1}(-1)^{i-k}{i \choose k-1}\beta_i \right]m^{k-1}  \end{array}$$
If Equation (\ref{pformu4}) holds for any $m>0$, we have that $\sum_{i=k}^{d+1}(-1)^{i-k}{i \choose k-1}\beta_i=\gamma_k$ holds for any $k=1,\cdots,d+1$. Let
$$A=\left(
  \begin{array}{cccc}
    a_{11} & a_{12} & \cdots & a_{1,d+1} \\
    0 & a_{22} & \cdots  & a_{2,d+1}  \\
       \vdots& \vdots  & \ddots  & \vdots  \\
     0 & 0  & \cdots&  a_{d+1,d+1} \\
  \end{array}
\right)=\left(
  \begin{array}{cccc}
     {1\choose 0} & -{2\choose 0} & \cdots & (-1)^{d }{d+1 \choose 0} \\
    0 & 2 \choose 1 & \cdots  &  (-1)^{d-1}{d+1 \choose 1}  \\
      \vdots &  \vdots &  \ddots  & \vdots  \\
     0 & 0  & \cdots&  {d+1 \choose d} \\
  \end{array}
\right)
$$
and $\beta=(\beta_1,\cdots,\beta_{d+1})^T$, and $\gamma=(\gamma_1,\cdots,\gamma_{d+1})^T$.
We have
$$ A\beta=\gamma. $$
It is easy to verify  that $\beta$ in Theorem \ref{solution1} is the solution of $ A\beta=\gamma $. \hfill$\blacksquare$

\vspace{0.5cm}
 \textbf{{Proof of Corollary \ref{isolated}}}

Let  $v_1,\cdots,v_j$ be the $j$ isolated vertices, $v^\prime_1,\cdots,v^\prime_m$ be the m extended vertices. $V({\cal K}_1)$ be the vertices in ${\cal K}_1$, $V({\cal K})$ be the vertices in ${\cal K}$. Clearly, we have $V({\cal K})=V({\cal K}_1)\cup \{v_1,\cdots,v_j\}$

 Because $\cal K$ is  composed of    ${\cal K}_1$ and $j$ isolated vertices, Equation (\ref{isonodes}) holds when $m=0$ since $\mbox{Size}({\cal K})= \mbox{Size}(({\cal K}_1))$.  Consider the case $m=1$.   From Theorem \ref{recursiveformula1}, we have

\begin{equation}\label{isonodesproof1}f\left({\cal K},{m}\right)=  m \cdot f\left({\cal K},{m-1}\right)+\sum_{v\in  V({\cal K}) } f\left( {\cal K}_{N_v},{m}\right)  \frac{\mbox{Size}\left({\cal K}^{(v)}\right)}{\mbox{Size} \left({\cal K}_{N_v}\right)}.\end{equation}

 Since  $m-1=0$, we have $f({\cal K},{m-1})=\mbox{Size}({\cal K})=\mbox{Size}({\cal K}_1)$, and $m \cdot f\left({\cal K},{m-1}\right)=m\cdot f({\cal K}_1)$. Moreover, for  any $v\in V({\cal K}_1)$,
  $\mbox{Size}({\cal K}^{(v)})=\mbox{Size}(({\cal K}_1)^{(v)})$  and $ {\cal K}_{N_v} = ({\cal K}_1)_{N_v} $  hold. For any $v\in \{v_1,\cdots,v_j\}$,  $\mbox{Size}({\cal K}^{(v)})=\mbox{Size}({\cal K}_1)$ and ${\cal K}_{N_v}$ is a null graph; it follows   $f({\cal K}_{N_v},m)=m!$ and $\mbox{Size}({\cal K}_{N_v})=1$. From Equation  (\ref{isonodesproof1}), we have that

$$\begin{array}{rl}f\left({\cal K},{m}\right)
&= m\cdot \mbox{Size}\left({\cal K}_1\right)+\sum_{v\in  V({\cal K}_1) } f\left( {({\cal K}_1)}_{N_v},{m}\right)  \frac{\mbox{Size} \left({\cal K}_1^{(v)}\right)}{\mbox{Size}\left(({{\cal K}_1})_{N_v}\right)}+\mbox{Size}({\cal K}_1)jm!  \\
&=f({\cal K}_1,{m})+\mbox{Size}({\cal K}_1)jm!=f({\cal K}_1,{1})+\mbox{Size}({\cal K}_1)j \end{array} $$

We have Equation (\ref{isonodes}) holds for $m=1$.
Suppose that Equation (\ref{isonodes}) holds for $m=k-1$, consider $m=k$, from Equation  (\ref{isonodesproof1}), we have
\begin{equation}\label{proofa1}f\left({\cal K},{k}\right)=   k \cdot f\left({\cal K},{k-1}\right)+ \sum_{v\in  V({\cal K}_1) } f\left( {({\cal K}_1)}_{N_v},{k}\right)  \frac{ \mbox{Size} \left({\cal K}_1^{(v)}\right)}{\mbox{Size}\left(({{\cal K}_1})_{N_v}\right)}+\mbox{Size}({\cal K}_1)jk!
\end{equation}

Since Equation (\ref{isonodes}) holds for $m=k-1$, we have
\begin{equation}\label{proofa2} f\left({\cal K},{k-1}\right)= f\left({\cal K}_1,{k-1}\right) +j(k-1)\mbox{Size}({\cal K}_1)(k-1)!  .
\end{equation}
From Theorem \ref{recursiveformula1}, we can get
\begin{equation}\label{proofa3} f\left({\cal K}_1,{k}\right)=k f\left({\cal K}_1,{k-1}\right) + \sum_{v\in  V({\cal K}_1) } f\left( {({\cal K}_1)}_{N_v},{k}\right)  \frac{\mbox{Size}\left({\cal K}_1^{(v)}\right)}{\mbox{Size}\left(({{\cal K}_1})_{N_v}\right)}.
\end{equation}

From Equation (\ref{proofa1}),  (\ref{proofa2}), and (\ref{proofa3}), we have

$$\begin{array}{rl}f\left({\cal K},{k}\right)=&f\left( {\cal K}_1 ,{k}\right)+ j(k-1)\mbox{Size}({\cal K}_1)k! + \mbox{Size}({\cal K}_1)jk!\\
=&f\left({\cal K}_1,{k}\right)+ \mbox{Size}({\cal K}_1)jkk!. \end{array} $$

As a result, Equation (\ref{isonodes}) holds for any integer $m\geq 0.$ \hfill$\blacksquare$

\vspace{0.5cm}
 \textbf{{Proof of Corollary \ref{threetype}}}

\emph{
The proof of (1)}

When $\cal K$ is null graph,   ${\cal K}^{m+}$ is a completed graph with $m$ vertices,  the result (1)  holds obviously.

\emph{
The proof of (2)}

Let $d_1,d_2,d_3,\cdots,d_p$ be degrees of vertices  $v_1, \cdots,v_p$  in $\cal K$,   we have $\sum_{i=1}^p d_i=2(p-1)$.
Consider $g({\cal K},m)$ defined in Equation (\ref{iterizeformu2}),

$$g({\cal K},m)=\sum_{i=1 }^p \frac{f ({\cal K}_{N_{v_i}},{m})}{m!}  \frac{\mbox{Size} ({\cal K}^{(v_i)})}{\mbox{Size} ({\cal K}_{N_{v_i}})}.$$

Since $\cal K$ is a tree, we have that ${\cal K}_{N_{v_i}}$ is composed of $d_i$ isolated vertices, so
$\frac{f ({\cal K}_{N_v},{m})}{m!}=1+d_im$. We also have  ${f ({\cal K}^{(v_i)})=1}$ and ${f ({\cal K}_{N_{v_i}})}=1$ if  $\cal K$ is a tree. Consequently,
$$g({\cal K},m)=\sum_{i=1}^p (1+d_im)=p+2(p-1)m$$
The result (2) holds  according to Theorem \ref{solution1}.

\emph{
The proof of (3)}

 Consider a   tree plus graph, if it is chordal, the added edge must be in a triangle, otherwise, the tree plus graph is not chordal.
Let $d_1,d_2,d_3,\cdots,d_p$ be degrees of vertices $v_1, \cdots,v_p$   in $\cal K$ and $d_1,d_2,d_3$  are the degrees of the three vertices in the triangle,  we have $\sum_{i=1}^p d_i=2p$. Moreover,considering the induced subgraph of $\cal K$ over   $N_{v_i}$, we have that  ${\cal K}_{N_{v_i}}$ is composed of an edge and $d_i-2$ isolated vertices for $i=1,2,3$, and ${\cal K}_{N_{v_i}}$ just contains  $d_i$ isolated vertices for $i=4,5,\cdots,p$.    Following Corollary \ref{isolated}, we can calculate $g({\cal K},m)$ as following
$$\begin{array}{rl}g({\cal K},m)=&\frac{1}{m!}\left[2(d_1+d_2+d_3-6)mm!+3(m+2)!+2\sum_{i\neq 1,2,3}(d_imm!+m!)\right]\\
=&3m^2+4pm-3m+2p \end{array}    $$
The result (3) holds  according to Theorem \ref{solution1}.

\emph{
The proof of (4)}

Consider a vertex $v$ in $\cal K$, we have that $({\cal K}^{m+})^{(v)}$ contains $p/2$ chain components, in which one is a completed graph with $m+1$ vertices, and the others are one-edge graphs.
We can calculate $g({\cal K},m)$ defined in   Equation (\ref{incrementg}) as following

$$g({\cal K},m)=2^{p/2}pm/2+2^{p/2}p/2.$$
As a result, the result (4)   holds according to Theorem \ref{solution1}. \hfill$\blacksquare$

\vspace{0.5cm}
 \textbf{{Proof of Corollary \ref{countgm}}}

According to the definition of $g({\cal K},m)$ in Equation (\ref{iterizeformu2})
$$ g({\cal K},m)=\sum_{j=1}^J\sum_{v\in V({\cal K}_j)} \frac{f ({\cal K}_{N_v},{m})}{m!}  \frac{\mbox{Size} ({\cal K}^{(v)})}{\mbox{Size} ({\cal K}_{N_v})}.$$

Because $\cal K$ is composed of  ${\cal K}_1,\cdots,{\cal K}_J$ that are $ J $ isolated connected graphs, we have that ${\cal K}_{N_v}={\cal K}_{j,N_v} $, and  $\mbox{Size}({\cal K}^{(v)})=\mbox{Size}({\cal K}_j^{(v)})\prod_{l\neq j} \mbox{Size}({\cal K}_l)$ $=\mbox{Size}({\cal K}_j^{(v)}) \frac{\mbox{Size}({\cal K})}{\mbox{Size}({\cal K}_j)}$. Consequently, Corollary \ref{countgm} holds. \hfill$\blacksquare$

\bibliography{reference}

\begin{thebibliography}{13}
\providecommand{\natexlab}[1]{#1}
\providecommand{\url}[1]{\texttt{#1}}
\expandafter\ifx\csname urlstyle\endcsname\relax
  \providecommand{\doi}[1]{doi: #1}\else
  \providecommand{\doi}{doi: \begingroup \urlstyle{rm}\Url}\fi

\bibitem[Andersson et~al.(1997)Andersson, Madigan, and
  Perlman]{andersson1997characterization}
S.~A. Andersson, D.~Madigan, and M.~D. Perlman.
\newblock {A characterization of Markov equivalence classes for acyclic
  digraphs}.
\newblock \emph{The Annals of Statistics}, 25\penalty0 (2):\penalty0 505--541,
  1997.

\bibitem[Chickering(2002)]{chickering2002learning}
D.~M. Chickering.
\newblock {Learning equivalence classes of Bayesian-network structures}.
\newblock \emph{The Journal of Machine Learning Research}, 2:\penalty0
  445--498, 2002.

\bibitem[Gillispie(2006)]{gillispie2006formulas}
S.~B. Gillispie.
\newblock {Formulas for counting acyclic digraph Markov equivalence classes}.
\newblock \emph{Journal of Statistical Planning and Inference}, 136\penalty0
  (4):\penalty0 1410--1432, 2006.

\bibitem[Gillispie and Perlman(2002)]{gillispie2002size}
S.B. Gillispie and M.D. Perlman.
\newblock {The size distribution for Markov equivalence classes of acyclic
  digraph models}.
\newblock \emph{Artificial Intelligence}, 141\penalty0 (1-2):\penalty0
  137--155, 2002.

\bibitem[He and Geng(2008)]{he2008active}
Yangbo He and Zhi Geng.
\newblock Active learning of causal networks with intervention experiments and
  optimal designs.
\newblock \emph{Journal of Machine Learning Research}, 9:\penalty0 2523--2547,
  2008.

\bibitem[He et~al.(2015)He, Jia, and Yu]{he2015counting}
Yangbo He, Jinzhu Jia, and Bin Yu.
\newblock Counting and exploring sizes of markov equivalence classes of
  directed acyclic graphs.
\newblock \emph{Journal of Machine Learning Research}, 16:\penalty0 2589--2609,
  2015.

\bibitem[Maathuis et~al.(2009)Maathuis, Kalisch, and
  B{\"u}hlmann]{maathuis2009estimating}
M.~H. Maathuis, M.~Kalisch, and P.~B{\"u}hlmann.
\newblock {Estimating high-dimensional intervention effects from observational
  data}.
\newblock \emph{The Annals of Statistics}, 37\penalty0 (6A):\penalty0
  3133--3164, 2009.
\newblock ISSN 0090-5364.

\bibitem[Pearl(2000)]{pearl2000causality}
J.~Pearl.
\newblock \emph{{Causality: Models, Reasoning, and Inference}}.
\newblock Cambridge Univ Pr, 2000.

\bibitem[Robinson(1973)]{robinson1973counting}
R.~Robinson.
\newblock {Counting labeled acyclic digraphs}.
\newblock \emph{New Directions in the Theory of Graphs}, pages 239--273, 1973.

\bibitem[Robinson(1977)]{robinson1977counting}
R.~Robinson.
\newblock Counting unlabeled acyclic digraphs.
\newblock In \emph{Combinatorial mathematics V}, pages 28--43. Springer, 1977.

\bibitem[Spirtes et~al.(2001)Spirtes, Glymour, and
  Scheines]{spirtes2001causation}
P.~Spirtes, C.N. Glymour, and R.~Scheines.
\newblock \emph{{Causation, Prediction, and Search}}.
\newblock The MIT Press, 2001.

\bibitem[Steinsky(2003)]{steinsky2003enumeration}
B.~Steinsky.
\newblock {Enumeration of labelled chain graphs and labelled essential directed
  acyclic graphs}.
\newblock \emph{Discrete mathematics}, 270\penalty0 (1-3):\penalty0 266--277,
  2003.

\bibitem[Verma and Pearl(1990)]{verma1990equivalence}
T.~Verma and J.~Pearl.
\newblock {Equivalence and synthesis of causal models}.
\newblock In \emph{Proceedings of the Sixth Annual Conference on Uncertainty in
  Artificial Intelligence}, page 270. Elsevier Science Inc., 1990.

\end{thebibliography}

\end{document}